\documentclass{llncs}
\usepackage{emily}
\title{Layered controller synthesis for dynamic multi-agent systems\thanks{This work was partially funded by ANR project TickTac (ANR-18-CE40-0015).}}

\author{Emily Clement\inst{1}
\and Nicolas Perrin-Gilbert \inst{1}
\and Philipp Schlehuber-Caissier \inst{2}
 }
\authorrunning{F. Author et al.}

\institute{
    Sorbonne Université, CNRS, Institut des Systèmes Intelligents et de Robotique, ISIR, F-75005 Paris, France \email{lastname@sorbonne-universite.fr}
    \and EPITA Research Laboratory \email{firstname@lrde.epita.fr}}

\usepackage[backend=biber]{biblatex}
\begin{document}

\maketitle
\begin{abstract}
    In this paper we present a layered approach for multi-agent control problem, decomposed into three stages, each building upon the results of the previous one.
First, a high-level plan for a coarse abstraction of the system is computed,
relying on parametric timed automata augmented with stopwatches as they allow to efficiently model simplified dynamics of such systems.
In the second stage, the high-level plan, based on SMT-formulation,
mainly handles the combinatorial aspects of the problem,
provides a more dynamically accurate solution.
These stages are collectively referred to as the SWA-SMT solver.
They are correct by construction but lack a crucial feature:
they cannot be executed in real time.
To overcome this, we use SWA-SMT solutions as the initial training dataset for our last stage,
which aims at obtaining a neural network control policy.
We use reinforcement learning to train the policy,
and show that the initial dataset is crucial for the overall success of the method.

\end{abstract}
\section{Introduction}
\label{sec:Intro}

Controlling a system involving multiple agents sharing a common task
is a problem occurring in several domains such as mobile or industrial robotics.
Concrete instances range from controlling swarms of drones, autonomous vehicles
or warehouse robots.
The problem is studied for specific instances but
remains a difficult problem in general, especially in safety critical cases.
The main complexity stems from the different types of decisions to take:
such control problems often have a strong combinatorial side while the approach
also has to deal with the physical reality of the agents, whose state typically
evolves according to some differential equation, and limitations on the control
inputs have to be taken into account.
Finally, the controller has to be executed in real-time, which typically
limits the applicability of formal methods due to their high complexity.

In this paper, we propose a layered approach for synthesizing control strategies
for multi-agent dynamical systems, whose effectiveness we demonstrate
on an example of centralized traffic guidance used for illustration
throughout the paper.

The layered approach involves three stages, each one addressing a specific
aspect of the control problem by building on the results of the previous stage.
The first stage deals with the combinatorial side of the control problem:
using a sufficiently coarse abstraction of the system dynamics, one can rely on
timed automata augmented with stopwatches as a model.
Efficient algorithms exist to explore such models allowing us to find
a high-level plan that guarantees success in this abstract
setting.
The second stage takes the high-level plan as an input and refines it
using a more realistic model of the system while maintaining a high
degree of similarity between the refined and high-level solution.
In our running example, we formulate this as an SMT problem,
respecting the discrete version of the differential equation describing the system
while also taking into account the input and state constraints.
The complexity of this stage remains reasonable as the combinatorial
aspects have already been solved.
The final stage addresses the issue of real-time execution and generalization.
To this end, we train a neural network policy via reinforcement learning.
We use the two first stages to construct a dataset of successful episodes on many random instances of the problem, and exploit this dataset to guide the reinforcement learning towards good solutions. On our running example, we show that the initial dataset of solutions is crucial for the overall success: the reinforcement learning only succeeds if it has access to it.

The rest of the paper is structured as follows.
In section \ref{sec:Appli} we present our running example,
section \ref{sec:SoA}
briefly discusses related work, and then we describe each stage of the approach in its own section along with the necessary technical background: the first stage using timed automata in
section \ref{sec:ta}, the second SMT-based stage in section \ref{sec:smt}
and finally the synthesis of the actual controller based on
reinforcement learning in section \ref{sec:RL}.

\section{Running example: centralized traffic control}
\label{sec:Appli}
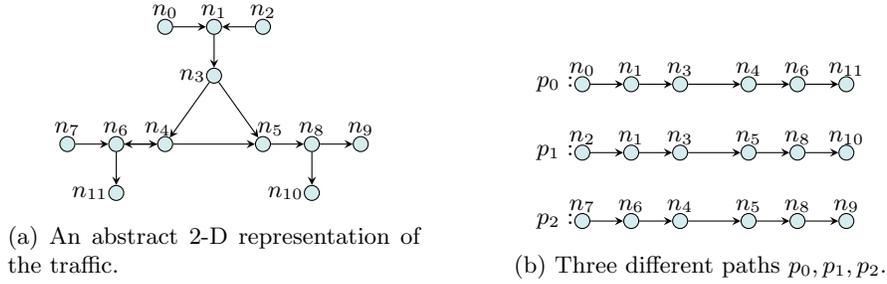
\begin{figure*}[htb]
    \tikzstyle{r-node}= [circle,fill,inner sep=2pt, color=middleSteelblue, draw=black]
    \begin{subfigure}[b]{0.45\linewidth}
        \centering
        \begin{tikzpicture}[scale=1.3,every node/.style={scale=1}]
            \node[r-node] (n-0) at (1.5,0) {};
            \node[r-node] (n-1) at (2,0) {}; % NOEUD DU MILIEU
            \node[r-node] (n-2) at (2.5,0) {};

            \node[r-node] (n-3) at (2,-0.5) {};
            \node[r-node] (n-4) at (1.5,-1.2) {};
            \node[r-node] (n-5) at (2.5,-1.2) {};

            \node[r-node] (n-6) at (1,-1.2) {};
            \node[r-node] (n-7) at (0.5,-1.2) {};
            \node[r-node] (n-8) at (3,-1.2) {};
            \node[r-node] (n-9) at (3.5,-1.2) {};
            \node[r-node] (n-10) at (3,-1.7) {};
            \node[r-node] (n-11) at (1,-1.7) {};

            \node[above] at (n-0) {$n_0$};
            \node[above] at (n-1) {$n_1$};
            \node[above] at (n-2) {$n_2$};
            \node[left] at (n-3) {$n_3$};
            \node[above, xshift=-0.1cm] at (n-4) {$n_4$};
            \node[above, xshift=0.1cm] at (n-5) {$n_5$};
            \node[above] at (n-6) {$n_6$};
            \node[above] at (n-7) {$n_7$};
            \node[above] at (n-8) {$n_8$};
            \node[above] at (n-9) {$n_9$};
            \node[left] at (n-10) {$n_{10}$};
            \node[left] at (n-11) {$n_{11}$};

            % LEFT PATH
            \path[thin-fleche]
                (n-0) edge (n-1)
                (n-1) edge (n-3)
                (n-3) edge (n-4)
                (n-4) edge (n-6)
                (n-6) edge (n-11);
            
            \path[thin-fleche]
                (n-2) edge (n-1)
                % useless (n-1) edge (n-3)
                (n-3) edge (n-5)
                (n-5) edge (n-8)
                (n-8) edge (n-10);

            \path[thin-fleche] 
                (n-7) edge (n-6)
                (n-6) edge (n-4)
                (n-4) edge (n-5)
                (n-8) edge (n-9);
        \end{tikzpicture}
        \subcaption{An abstract $2$-D representation of the traffic.}
        \label{fig:running-ex-2d}
    \end{subfigure} 
    \hfill
    \begin{subfigure}[b]{0.49\linewidth}
        \centering 
        \begin{tikzpicture}[scale=1.3,every node/.style={scale=1}]

            \node[r-node] (n-0-0) at (0,0) {};
            \node[r-node] (n-1-0) at (0.5,0) {}; % NOEUD DU MILIEU
            \node[r-node] (n-3-0) at (1,0) {};
            \node[r-node] (n-4-0) at (1.7,0) {};
            \node[r-node] (n-6-0) at (2.2,0) {};
            \node[r-node] (n-11-0) at (2.7,0) {};

            \node[left] at (n-0-0) {$ \dPath_{0}:$};
            \node[above] at (n-0-0) {$n_0$};
            \node[above] at (n-1-0) {$n_1$};
            \node[above] at (n-3-0) {$n_3$};
            \node[above] at (n-4-0) {$n_4$};
            \node[above] at (n-6-0) {$n_6$};
            \node[above] at (n-11-0) {$n_{11}$};

            % LEFT PATH
            \path[thin-fleche]
                (n-0-0) edge (n-1-0)
                (n-1-0) edge (n-3-0)
                (n-3-0) edge (n-4-0)
                (n-4-0) edge (n-6-0)
                (n-6-0) edge (n-11-0);

            \node[r-node] (n-2-1) at (0,-0.7) {};
            \node[r-node] (n-1-1) at (0.5,-0.7) {}; % NOEUD DU MILIEU
            \node[r-node] (n-3-1) at (1,-0.7) {};
            \node[r-node] (n-5-1) at (1.7,-0.7) {};
            \node[r-node] (n-8-1) at (2.2,-0.7) {};
            \node[r-node] (n-10-1) at (2.7,-0.7) {};

            \node[left] at (n-2-1) {$ \dPath_{1}:$};
            \node[above] at (n-1-1) {$n_1$};
            \node[above] at (n-2-1) {$n_2$};
            \node[above] at (n-3-1) {$n_3$};
            \node[above,] at (n-5-1) {$n_5$};
            \node[above] at (n-8-1) {$n_8$};
            \node[above] at (n-10-1) {$n_{10}$};

            \path[thin-fleche]
                (n-2-1) edge (n-1-1)
                (n-1-1) edge (n-3-1)
                (n-3-1) edge (n-5-1)
                (n-5-1) edge (n-8-1)
                (n-8-1) edge (n-10-1);

            \node[r-node] (n-7-2) at (0,-1.4) {};
            \node[r-node] (n-6-2) at (0.5,-1.4) {}; % NOEUD DU MILIEU
            \node[r-node] (n-4-2) at (1,-1.4) {};
            \node[r-node] (n-5-2) at (1.7,-1.4) {};
            \node[r-node] (n-8-2) at (2.2,-1.4) {};
            \node[r-node] (n-9-2) at (2.7,-1.4) {};

            \node[left] at (n-7-2) {$ \dPath_{2}:$};
            \node[above] at (n-4-2) {$n_4$};
            \node[above] at (n-5-2) {$n_5$};
            \node[above] at (n-6-2) {$n_6$};
            \node[above] at (n-7-2) {$n_7$};
            \node[above] at (n-8-2) {$n_8$};
            \node[above] at (n-9-2) {$n_9$};

            \path[thin-fleche] 
                (n-7-2) edge (n-6-2)
                (n-6-2) edge (n-4-2)
                (n-4-2) edge (n-5-2)
                (n-5-2) edge (n-8-2)
                (n-8-2) edge (n-9-2);
        \end{tikzpicture}
        \subcaption{Three different paths $ \dPath_{0}, \dPath_{1}, \dPath_{2} $.}
        \label{fig:running-ex-three-path}
    \end{subfigure}
    \caption{An abstract representation of our running example, arrows indicate which direction can be taken.}
    \label{fig:running-ex}
\end{figure*}

Let us first present a multi-agent system used as running example
(fig \ref{fig:running-ex-2d})
to illustrate our method throughout the article.
In this example, each agent models a physical car on a road network.
Each of the cars is given a fixed path to follow and it needs to attain its
designated goal position from its initial position, while maintaining a security distance
to the other cars.
In such a setting the dynamics can be reduced to a second order ordinary differential equation
with lower and upper bounds on the first and second derivative (corresponding to the speed
and acceleration of the car).
\subsection{Multi-agent traffic}
We model the traffic as a network of sections on which a fixed number of cars have to navigate.
Cars drive along their \textit{paths} which consist of a list of sections.
Cars cannot overtake or cross each other on the same section. A minimum security distance must be maintained at all time.

To define a \textbf{section}, denoted $ \sect $,
we specify its beginning node, $ \node{b} $, its end node, $ \node{e} $, and its length $ \sectL $.
We can therefore write a section $ \sect := \sectParam{\node{b}}{\node{e}}{\sectL} $.
To specify its direction ($ \dirTrue $ or $ \dirFalse $),
we augment a section $ \sect $ with a direction $ \direction \in \{ \dirTrue, \dirFalse \} $ into a \textbf{directed section}, denoted $ \dir{\sect} $, or $ ( \sect, \direction ) $.
The notion of beginning and end nodes
 of a section is extended to directed sections, reversing the two nodes if $ \direction = \dirFalse $.
A directed section $ \dir{\sect'} $ is a \textbf{successor} (\resp \textbf{predecessor}) of the directed section $ \dir{\sect} $ if
 $ \node{e} = \node{b}' $ (\resp $ \node{e}' = \node{b} $).
The sections $ \dir{\sect} $ and $ \dir{\sect'} $ are said to be \textbf{neighbours} if $ \dir{\sect} $
is either a \textbf{successor} or \textbf{predecessor} of $ \dir{\sect'} $.

A \textbf{path} $ \dPath $ is defined as a finite list of directed sections:
$ \dPath = ( \dir{\sect_k})_{ 0 \leq k \leq m} $ such that
for all $ k \in [ 0, m - 1] $, $ \dir{\sect_{k+1}} $ is a successor of $ \dir{\sect_{k}} $.
The end (\resp beginning) node of a path is the end (\resp beginning) node of its last (\resp first) directed section.
By abuse of notation, we denote $ \dir{\sect} \in \dPath $ if there exists an index $ k $ such that
$ \dir{\sect} = \dir{\sect_k} $.

\textit{In fig \ref{fig:running-ex-three-path}, $3$ paths are described: directed sections $ (\sectParam{ \node{4} }{ \node{6} }{ \sectL }, \dirFalse ) $ and
$ (\sectParam{ \node{4} }{ \node{5} }{ \sectL }, \dirTrue ) $ are neighbours, but
$ (\sectParam{ \node{4} }{ \node{6} }{ \sectL }, \dirTrue ) $ and
$ (\sectParam{ \node{4} }{ \node{5} }{ \sectL }, \dirTrue ) $ are not.}

A \textbf{car} is defined as a tuple of an index $ i $ and a path $ \dPath $.
A \textbf{car traffic}, denoted $ \setOfCars $, is a set of cars.
The set of sections (\resp directed sections) of the cars of a car traffic is denoted $ \setOfSects $ (\resp $ \setOfDSects $).
A section $ \sect $ is an \textbf{intersection} if there exist
 two different paths $ \dPath, \dPath' $ and
 two directions $ \direction, \direction' \in \{ \dirTrue, \dirFalse \}$
 such that $ ( \sect, \direction) \in \dPath, ( \sect, \direction' ) \in \dPath' $.

\textit{Let us illustrate again with our example of fig \ref{fig:running-ex}
with a car traffic composed of three cars $ (i, \dPath_{i})_{i = 0,1,2}$.
Intersections here are
$\sectParam{ \node{1} }{ \node{3} }{ \sectL }$,
$\sectParam{ \node{4} }{ \node{6} }{ \sectL }$ and
$\sectParam{ \node{5} }{ \node{8} }{ \sectL }$.}

Since the path of each car is fixed, we only need to keep track of its speed and its progress along its path.

\subsection{Collision avoidance problem}
Given a security distance $ \sectDist $, the initial positions of all cars, bounds on their speed, acceleration and deceleration,
the goal is to find trajectories for all cars, which can be interpreted as a \textbf{centralized strategy},
that respect the three following rules.

Let $ \sectTimed{ \car }{ t } $ (\resp $ \sectdTimed{ \car }{ t } $) denote the current section
(\resp directed section) of car $ \car $ at time $t$ and $ \relativePosition{ \car }{ t }{ \dir{\sect} } $ its current relative position within $\dir{\sect}$.

1. \textbf{Same directed section}: 
for all cars $ \car_i, \car_j \in \setOfCars $, $\car_i \neq \car_j$,
for all $ t \geq 0 $,
if $ \sectdTimed{ \car }{ t } = \sectdTimed{ \car ' }{ t } = \dir{\sect} $
 then:
    %\begin{equation}
    $
        \vert \relativePosition{ \car_i }{ t }{ \dir{\sect} }
        -
        \relativePosition{ \car_j }{ t }{ \dir{\sect} } \vert
        \geq \sectDist
    $
    %\end{equation}

2. \textbf{Neighbouring sections}: If there exists a section $ \sect' $ such that there exists two cars $ \car_i = (i,[ \cdots , (\sect', \direction_i'), \cdots ]) \in \setOfCars $,
and $ \car_j = (j,[ \cdots , (\sect, \direction_j), (\sect', \direction_j'), (\sect'', \direction_j''), \cdots ]) \in \setOfCars $ sharing the section $ \sect' $,
then:
\begin{itemize}
\item If $\direction_i' = \direction_j'$, then for all $t$, such that $ \sectdTimed{ \car_i }{ t } = (\sect', \direction_i') $
, $ \sectdTimed{ \car_j }{ t } = (\sect'', \direction_j'') $, we have
    %\begin{equation}
    $
        \sectL' - \relativePosition{ \car_i }{ t }{ (\sect', \direction_i') }
        +
        \relativePosition{ \car_j }{ t }{ (\sect'', \direction_j'') }
        \geq \sectDist
    $.
    %\end{equation}
\item If $\direction_i' \neq \direction_j'$, then for all $t$, such that $ \sectdTimed{ \car_i }{ t } = (\sect', \direction_i') $
    , $ \sectdTimed{ \car_j }{ t } = (\sect, \direction_j) $, we have
        %\begin{equation}
        $
            \sectL' - \relativePosition{ \car_i }{ t }{ (\sect', \direction_i') }
            +
            \sectL - \relativePosition{ \car_j }{ t }{ (\sect, \direction_i) }
            \geq \sectDist
        $
        %\end{equation}
\end{itemize}

3. \textbf{Same section, opposite direction}: 
for all section $ \sect \in \setOfSects $,
for all $ t \geq 0 $
 and
 for each pair of cars $ \car_i, \car_j \in \setOfCars $:
    %\begin{equation}
    $
        \lnot
        (
            \sectdTimed{ \car_i }{ t } = ( \sect, \dirTrue )
            \wedge
            \sectdTimed{ \car_j }{ t } = ( \sect, \dirFalse )
        )
    $
    %\end{equation}

\subsection{Running example}

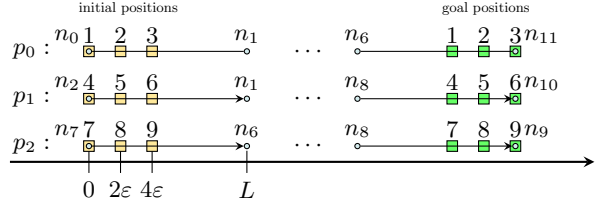
\begin{figure*}[htb]
    \centering
    \tikzstyle{r-node}= [circle,fill,inner sep=0.7pt, color=middleSteelblue, draw=black]
    \tikzstyle{car-pos} = [rectangle, inner xsep=2pt, inner ysep= 2pt, color=middleRed, draw=black]
    \tikzstyle{car-pos-init} = [car-pos, fill = deepMustard]
    \tikzstyle{car-pos-goal} = [car-pos, fill = deepGreen]
    \begin{tikzpicture}[scale=2.1,every node/.style={scale=1}]

        % \draw[loosely dotted] (0,-1) grid (3,0);
        \draw[fleche] (-0.5,-0.7) -- (3.2,-0.7) node[right] {};
        \foreach \x/\xtext in {
            0/0,
            0.2/2\sectDist,
            0.4/4\sectDist,
            1/\sectL}
        \draw[shift={(\x,-0.7)}] (0pt,2pt) -- (0pt,-2pt) node[below] {$\xtext$};

        % CARS INITIAL POSITION
        \node[car-pos-init] (c-1-0) at (0, 0) {};
        \node[above] at (c-1-0) {$1$};

        \node[car-pos-init] (c-2-0) at (0.2, 0) {};
        \node[above] at (c-2-0) {$2$};

        \node[car-pos-init] (c-3-0) at (0.4, 0) {};
        \node[above] at (c-3-0) {$3$};

        \node[car-pos-init] (c-4-0) at (0, -0.3) {};
        \node[above] at (c-4-0) {$4$};

        \node[car-pos-init] (c-5-0) at (0.2, -0.3) {};
        \node[above] at (c-5-0) {$5$};

        \node[car-pos-init] (c-6-0) at (0.4, -0.3) {};
        \node[above] at (c-6-0) {$6$};

        \node[car-pos-init] (c-7-0) at (0, -0.6) {};
        \node[above] at (c-7-0) {$7$};

        \node[car-pos-init] (c-8-0) at (0.2, -0.6) {};
        \node[above] at (c-8-0) {$8$};

        \node[car-pos-init] (c-9-0) at (0.4, -0.6) {};
        \node[above] at (c-9-0) {$9$};

        % CAR FINAL POSITION
        \node[car-pos-goal] (c-1-1) at (2.3, 0) {};
        \node[above] at (c-1-1) {$1$};

        \node[car-pos-goal] (c-2-1) at (2.5, 0) {};
        \node[above] at (c-2-1) {$2$};

        \node[car-pos-goal] (c-3-1) at (2.7, 0) {};
        \node[above] at (c-3-1) {$3$};

        \node[car-pos-goal] (c-4-1) at (2.3, -0.3) {};
        \node[above] at (c-4-1) {$4$};

        \node[car-pos-goal] (c-5-1) at (2.5, -0.3) {};
        \node[above] at (c-5-1) {$5$};

        \node[car-pos-goal] (c-6-1) at (2.7, -0.3) {};
        \node[above] at (c-6-1) {$6$};

        \node[car-pos-goal] (c-7-1) at (2.3, -0.6) {};
        \node[above] at (c-7-1) {$7$};

        \node[car-pos-goal] (c-8-1) at (2.5, -0.6) {};
        \node[above] at (c-8-1) {$8$};

        \node[car-pos-goal] (c-9-1) at (2.7, -0.6) {};
        \node[above] at (c-9-1) {$9$};

        \node[r-node] (n-0-0) at (0,0) {};
        \node[r-node] (n-1-0) at (1,0) {}; % NOEUD DU MILIEU
        \node[r-node] (n-6-0) at (1.7,0) {};
        \node[r-node] (n-11-0) at (2.7,0) {};

        \draw (1.4, 0) node[]{$ \cdots $};
        \node[above left] at (n-0-0) {$\node{0}$};
        \node[above] at (n-1-0) {$\node{1}$};
        \node[above] at (n-6-0) {$\node{6}$};
        \node[above right] at (n-11-0) {$\node{11}$};

        % LEFT PATH
        \path[-]
            (n-0-0) edge (n-1-0)
            % (n-1-0) edge (n-3-0)
            % (n-3-0) edge (n-4-0)
            % (n-4-0) edge (n-6-0)
            (n-6-0) edge (n-11-0);

        \node[r-node] (n-2-1) at (0,-0.3) {};
        \node[r-node] (n-1-1) at (1,-0.3) {}; % NOEUD DU MILIEU
        \node[r-node] (n-8-1) at (1.7,-0.3) {};
        \node[r-node] (n-10-1) at (2.7,-0.3) {};

        \node[left, xshift = -0.4cm] at (n-0-0) {$ \dPath_{0}:$};
        \node[left, xshift = -0.4cm] at (n-2-1) {$ \dPath_{1}:$};
        \node[above] at (n-1-1) {$\node{1}$};
        \draw (1.4, -0.3) node[]{$ \cdots $};
        \node[above left] at (n-2-1) {$\node{2}$};
        \node[above] at (n-8-1) {$\node{8}$};
        \node[above right] at (n-10-1) {$\node{10}$};

        \path[thin-fleche]
            (n-2-1) edge (n-1-1)
            (n-8-1) edge (n-10-1);

        \node[r-node] (n-7-2) at (0,-0.6) {};
        \node[r-node] (n-6-2) at (1,-0.6){}; % NOEUD DU MILIEU
        \node[r-node] (n-8-2) at (1.7,-0.6) {};
        \node[r-node] (n-9-2) at (2.7,-0.6) {};

        \node[left, xshift=-0.4cm] at (n-7-2) {$ \dPath_{2}:$};
        \draw (1.4, -0.6) node[]{$ \cdots $};
        \node[above left] at (n-7-2) {$\node{7}$};
        \node[above] at (n-6-2) {$\node{6}$};

        \node[above] at (n-8-2) {$\node{8}$};
        \node[above right] at (n-9-2) {$\node{9}$};

        \path[thin-fleche]
            (n-7-2) edge (n-6-2)
            (n-8-2) edge (n-9-2);

        \draw (-0.05,0.2) node[above right, xshift=-0.1cm, scale=0.6]{initial positions};
        \draw (2.25,0.2) node[above right, xshift=-0.1cm, scale=0.6]{goal positions};       
    \end{tikzpicture}
    \caption{Initial and goal position of cars}
    \label{fig:initial-pos-car}
\end{figure*}

Let us consider three possible paths, illustrated in fig \ref{fig:running-ex}.
All sections have the same security distance $ \sectDist $ and the same length, $ \sectL = 30 $, except for those from $ \node{3} $ to $ \node{5} $ and $ \node{3} $ to $ \node{4} $ that have length $ 30 \sqrt{2} $.
The cars are defined as $(i, \dPath_{0})_{i=1,2,3}$, $(i, \dPath_{1})_{i=4,5,6}$ and $(i, \dPath_{2})_{i=7,8,9}$.
They all have different initial and goal positions, starting with a security distance $ 2\sectDist $ between them as shown in fig \ref{fig:initial-pos-car}.
For instance, the initial position of car $2$ is $ 2\sectDist $ to the right of $ \node{0} $ in direction of $ \node{1}$.
Its goal position is $ \sectL  - 2\sectDist $ to the right of $ \node{6} $. All other cars are setup similarly.
Let us precise that, to make the implementation of the car traffic more convenient, we created additional nodes dedicated to the initial and goal positions of each car, omitted here for clarity.
Therefore in our actual implementation, the section from $ \node{0} $ to $ \node{1} $ is subdivided with nodes $ \node{0}' $ and $ \node{0}''$ representing the actual initial positions of car $ 2 $ and $ 3 $.

In the next sections, we present our method and apply it to this problem with several levels of abstraction.
Firstly, in section \ref{sec:ta}, we conceive a high-level plan relying on timed automata,
assuming that the speed of cars is discretized to two values, a nominal value $ \speed $ and $ 0 $,
and that cars can switch between these speeds instantly.

We also assume that all vehicles must respect the same security distance on all sections
(extending to different speed or security distance between sections is trivial).
Secondly, in section \ref{sec:smt}, we refine this

abstract model by allowing for arbitrarily many speed levels between $ \speed $ and $ 0 $ and also by respecting the maximal values for
acceleration and deceleration. In section \ref{sec:RL}, there are discrete time steps, but the neural network policy outputs continuous values for the acceleration/deceleration.

\section{Related work}
\label{sec:SoA}
There exists a rich literature on multi-agent systems, including
path planning or controller synthesis.
Providing a general overview over these topics is beyond the scope of this paper,
however we give a brief overview of the ones matching our objective the best.
In our model, decisions are taken by a centralized control agent (the reference trajectories) with perfect knowledge
and then executed by all executing agents (the cars). 
For a good survey of topics related to multi-agent systems we refer to \cite{dorri2018multi}.

Path planning, collision avoidance and controller synthesis for multi-agent systems in general
as well as centralized traffic control allow for a rich variety of useful
abstraction which in turn leads to a large spectrum of techniques and concepts
that can be applied.
Approaches can be differentiated into different categories:
fully discretized approaches largely ignoring the underlying dynamics of the
system fall into the category of multi-agent path finding.
Here the problem boils down to a graph search on (very) large graphs, see \cite{stern2019multi, DBLP:phd/hal/Queffelec21}.
Multi-agent motion planning in contrast takes into account the underlying the dynamics
(possibly even uncertainty) and works over continuous domains as done in \cite{chen2021scalable}.
We are positioned in between these approaches: we consider 
the dynamics of the system, however the road system is fixed, so the planning
is over a finite domain.

Another way to distinguish approaches is the complexity of the specification.
If the target is known and the only goal is to avoid collision,
less formal approaches ensuring safety and success in practice without
a proof can provide good results as shown in \cite{FioriniS98, van2008reciprocal}.
More complex specifications are taken into account by works like \cite{Kress-GazitFP07}
and \cite{bogh2022distributed}.
In \cite{Kress-GazitFP07}, controllers for drones verifying temporal logic
specifications are synthesized given low-level controllers guaranteeing to
bring them from one region to another exist; collision between different
drones is ignored as they are supposed to fly at different altitudes.
In \cite{bogh2022distributed} controllers for a fleet of warehouse robots
that have to share resources to fulfill different task in a near optimal manner
are learned.
Our final layer shares some characteristics of \cite{van2008reciprocal}:
it does not provide formal guarantees but is executable in real-time.
With \cite{bogh2022distributed} we share the idea of a layered approach however
to achieve different goals. In their work one controller is learned for
resource distribution and another for path-planning. Our approach in contrast uses
layers to refine plans with different levels of abstractions.
In \cite{li2018policy}, temporal logic task specifications are translated into real-valued functions that can be used as reward signals to guide reinforcement learning.

Finally, there exist also many works tackling explicitly collision avoidance
in traffic scenarios like \cite{HilscherLOR11, HilscherLO13, HilscherS16, ColomboV12, LoosP11}.
The works of \cite{HilscherLOR11, HilscherLO13, HilscherS16} rely on discretization
and overapproximation, then proving collision freeness for a given set controllers
and maneuvers in an offline manner.
In \cite{ColomboV12} only intersections are treated and not the problem of
sharing a section while driving in the same direction.
\cite{LoosP11} avoids collision by finding an optimal scheduling for the
traffic lights, which is however only applicable in a very restricted scenario.

\section{High level planning}
\label{sec:ta}
Let us present the first level of our method, based on Timed Automata (TA),
a well established tool to model real-time systems with timing constraints using clocks.
Besides time, these clocks can also be used to model other quantities if they
behave somewhat similarly to time in an abstract sense.
In our running example, we assign to each car a clock that tracks its progress along its path.
This basic framework is not expressive enough to obtain a useful model and
we need to use several extensions.
We augment the TA with stopwatches to represent the agent at standstill or driving at nominal speed.
We rely on channels to ensure the collision avoidance and relative order between cars.
To obtain reachability in minimal time, we dedicate a clock to represent global time
(therefore never stopped nor reset), which is used as a parameter that does not appear
in the constraints.

\subsection{Timed Automata, stopwatches and channels}
\label{subsec:ta/defs}
In this section we recall standard TA semantics as well as the needed extensions.

\begin{definition}[\cite{AlurD94}] \label{def:timed-auto}
    A Timed Automaton (TA) $\automata $ can be defined by the tuple $( \locationSet, \initLoc, \clocksSet, \invSet, \alphabet, \transitionSet ) $,
     where $\locationSet$ is a finite set of locations,
    $\initLoc$ is the initial location, $ \clocksSet $ is a finite set of clocks,
    $\alphabet $ a finite alphabet,  $ \inv:  \locationSet \rightarrow \guardConstraint$ is the function of invariants of locations and $ \transitionSet \subseteq \locationSet \times \guardConstraint \times \alphabet \times 2^{\clocksSet} \times  \locationSet $ a set of
    transitions.
    
    There are two types of transitions with the following semantics:
    \begin{itemize}
    \setlength{\itemsep}{1pt}
    \setlength{\parskip}{0pt}
    \setlength{\parsep}{0pt} 
    \item Time elapsing move: $ ( \loc, \valuation ) \xrightarrow{ \delay } ( \loc, \valuation' )$: elapses some amount of time $ \delay $ by setting
    $ \valuation ' = \valuation + \delay $ and is only allowed if $\valuation \models \invSet( \loc )$ and $ \valuation ' \models \invSet( \loc ) $
    
    \item Discrete transition: $\simpleTrans{( \loc, \valuation )}{( \loc', \valuation' )}$ indicates a discrete
    transition.
    This is only possible if \textbf{(1)} $ \valuation \models \inv ( \loc ) $ and $\valuation' \models \inv ( \loc' ) $
    and \textbf{(2)} there exists a transition $ \transition := (\valuation, \guard, \act,  \reset, \loc') \in \transitionSet $
    such that $ \valuation \models \inv ( \loc ) $, $ \valuation \models \guard $, $ \valuation ' = \valPostReset{ \valuation }{ \reset }$ and $\valuation' \models \inv ( \loc ')$.
    We say that $ \transition $ is labelled by $ \act $.
    \end{itemize}
\end{definition}

Here, $ \valuation + \delay $ denotes the valuation $ \valuation_{\delay} $
such that for any clock $ \clock $ in the set of clocks, denoted $ \clocksSet $, $ \valuation_{\delay} ( \clock ) = \valuation(\clock) + \delay $
 and $ \valPostReset{ \valuation }{ \reset } $ the valuation $ \valuation_{\reset} $ such that for $ \valuation_{ \reset } ( \clock ) = 0 $ if $ \clock \in \reset $ and $ \valuation ( \clock ) $ otherwise.
$ \guardConstraint $ defines the set of clock-constraints by a conjunction of simple (in-) equalities:
$\wedge_i  \cOne \Join \constr{i} $ for some clock $ \cOne \in  \clocksSet $, some constant $\constr{i} \in \mathbb{N}$ and
$\Join \in \{ \le, <, =, >, \ge \}$.
We denote $ \valuation \models \guard $ to express that a valuation $ \valuation $ satisfies the constraint of $ \guard \in \guardConstraint$.

\textbf{Location based Stopwatch Timed Automata} (SWA),
presented in \cite{AlurCHHH95},
are an extension of TA where clocks can be \enquote{frozen} on locations.
More formally, it is a tuple $ \calA = ( \locationSet, \initLoc, \clocksSet, \invSet, \alphabet, \swSet, \transitionSet)$ with
    $\locationSet, \initLoc, \clocksSet, \invSet, \alphabet, \transitionSet$ have the same definition as in def \ref{def:timed-auto}
     and $ \swSet: \locationSet \rightarrow 2^{\clocksSet} $ is a function assigning to a location $ \loc $ the set of stopped clocks at $ \loc $.
    The definition for discrete transitions is the same as in def \ref{def:timed-auto}, however the time-elapsing transition changes.
    $\transParam{ ( \loc, \valuation )}{ \delay }{ ( \loc, \valuation' )} $:
    \textbf{(1)} elapses some amount of time $ \delay $ by setting
    $\valuation' (  \cOne ) =  \valuation (  \cOne ) \text{ if } \cOne \in \swSet ( \loc )$, $ \valuation (  \cOne ) + \delay$ otherwise

Reachability of SWA is undecidable in general however, as shown in \cite{KPV98},  there exist decidable fragments like \textbf{Initialized Stopwatch Timed Automata} (ISWA) for which deciding reachability remains in PSPACE as for TA.

\begin{definition}[\cite{KPV98}]
    \textbf{Initialized Stopwatch Timed Automata} a SWA $ \automata $ is a (ISWA) if for any transition $ \transition = ( \location ,  \guard , \act,  \reset,  \loc')$,
    if $ (
            \cOne \in \swSet ( \loc )
            \wedge
            \cOne \notin \swSet( \loc' ) ) $ or
            $
            (
                \cOne \notin \swSet ( \loc )
                \wedge
                \cOne \in \swSet ( \loc' )
            ) $, then $\cOne \in  \reset $.
\end{definition}

That is a clock is only started or stopped if it is also reset.
We will show in section \ref{subsec:ta/application} that our TA abstraction
of the running example falls into this category.

\begin{definition}[\cite{BrandZ83}]
    \textbf{Channel systems} are finite automata augmented with a finite number of channels.
    They can be thought of as FIFO (First In First Out) queues for symbols
    used for asynchronous communication.
    During a transition, we can either \textbf{(1)} Leave the channels untouched
    \textbf{(2)} Push a symbol into a channel $ \channels $, denoted $ \location \xrightarrow{\channels! \symbolLetter} \location '$
    indicating that the symbol $ \symbolLetter $ is appended to $\channels$.
    This is always possible if channels are unbounded, \ie can contain an unbounded
    number of symbols.
    \textbf{(3)} Peek and pop a symbol from a channel $\channels$, denoted $ \location \xrightarrow{\channels? \symbolLetter}  \location '$.
    This action looks at the head of $\channels$. If it contains the symbol $ \symbolLetter $, it is removed from $ \channels $ when taking the transition, otherwise the transition is deactivated.
\end{definition}

\input{sections/figures/cswa-ex.tex}

Bounded channel systems, that is channel systems in which channels can only contain a fixed number of symbols, are decidable.
They can be translated into a finite automaton (with exponentially many locations in both the number of different symbols and the maximal length of the channel), which can then form a synchronized product with the other automata.

As a high-level abstraction, we model our example as a parallel composition of ISWA communicating via strong
synchronization augmented by channels.

\subsection{Timed Automata representation of our running example}
\label{subsec:ta/application}

To model our car traffic with systems of Timed Automata, we suppose that each car begins/ends at the beginning/end node of a section and that it stops instantly.
Let us describe the automata we generate to model our system with a simple example. The full construction is detailed in appendix \ref{apdx:automata}.

\begin{figure}[]
    \centering
    \begin{tikzpicture}[->,>=stealth',auto,node distance=2.8cm, semithick]
        \node [state] (I) [] {$\arrived_{\dir{\sect}}$};
        \node [state] (A) [right of=I] {$\wait_{\dir{\sect'}}$};
        \node [state] (B) [right of=A] {$\driving_{\dir{\sect'}}$};
        \node [state] (C) [right of=B, xshift=0.5cm] {$\arrived_{\dir{\sect'}}$};
        \node [state] (Goal) [right of=C] {$\wait_{\dir{\sect''}}$};

        \node[] (invI) [below of = I, yshift=2.2cm]{$\{\}$};
        \node[] (invA) [below of = A, yshift=2.2cm]{$\{ \cOne_A \}$};
        \node[] (invB) [below of = B, yshift=2.2cm]{$\{\}$};
        \node[] (invC) [below of = C, yshift=2.2cm]{$\{\}$};
        \node[] (invG) [below of = Goal, yshift=2.2cm]{$\{ \cOne_A \}$};

        \path[fleche]
            (I) edge  [left, dashed] node[above] { $\begin{array}{l}  \cOne_A = \sectL_0 \end{array}$} node[below] {$\begin{array}{l}\sync_{\dir{\sect'}}(\cOne_A)\end{array}$}  (A)
            (A)  edge  [left] node[above] { $\begin{array}{l} \cOne_A = \sectL_0 \end{array}$} node[below] {$\begin{array}{l}\channels_{\dir{\sect'}} ? \cOne_A\end{array}$} (B)
            (B) edge  [left] node[above] { $\begin{array}{l} \cOne_A = \sectL_0 + \sectL \end{array}$} node[below] {$\begin{array}{l}\channels_{\dir{\sect''}} ! \cOne_A\end{array}$}  (C)
            (C) edge  [left, dashed] node[above] { $\begin{array}{l}  \cOne_A = \sectL_0 + \sectL \end{array}$} node[below] {$\begin{array}{l}\sync_{\dir{\sect''}}(\cOne_A)\end{array}$}  (Goal);
    \end{tikzpicture}
    \caption{The sub-automaton of our car Timed Automaton}
    \label{fig:ta-section}
\end{figure}
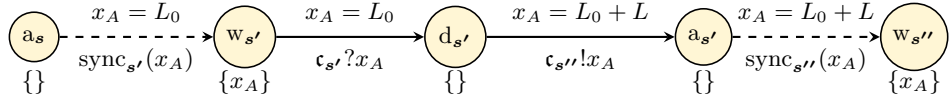

For each car $A$, the clock $ \cOne_A $ represents its progress along its path.
For each directed section $ \dir{\sect'} = (\sectParamP{ \node{i} }{ \node{j} }{ \sectL }, \direction) $ of its path,
$A$ performs three steps corresponding to three locations in the TA:
\textbf{(1)} waiting within the section $ \sect'$ (location $ \wait_{\dir{\sect'}}$) after entering in $ \sect'$ (action $\sync_{\dir{\sect'}}(\cOne_A)$); here the associated clock is stopped,
\textbf{(2)} driving in the section (location $ \driving_{\dir{\sect'}}$),
\textbf{(3)} arriving at the end of the section (location $ \arrived_{\dir{\sect'}}$) after having traveled a distance of $ \sectL $ (\resp letting $ \sectL $ time unit elapses).
We represent theses steps in fig \ref{fig:ta-section}, in which $\sectL_0$ denotes the accumulated distance to arrive at the end of section $\dir{\sect$}. The timed automaton of car $A$, is the concatenation of \enquote{sub-automata} of each directed section along the path.
If $\dir{\sect'}$ has no successor (\resp predecessor), $ \arrived_{\dir{\sect}}$ (\resp $\driving_{\dir{\sect'}}$) is the goal (\resp initial) location of the automaton.

\begin{figure}[]
    \centering
    \begin{tikzpicture}[->,>=stealth',auto,node distance=2cm, semithick]
        \node [initial state] (A) [] {$\free_{\sect}$};
        \node [state] (B) [right of=A, xshift=3cm] {$\blocked_{\sect, \dirTrue}$};
        \node [state] (C) [right of=B, xshift=3cm] {$\semifree_{\sect, \dirTrue}$};

        \path[fleche]
            (A)  edge [bend left=10] node[above, auto=left,yshift=0.05cm] {$\sync_{\dir{\sect}}(\cOne_{B})$} node[below] {$\cOne_{\sect} \leftarrow 0 $} (B)
            (A)  edge [bend right=30] node[above, auto=left, pos=0.7] {$\sync_{\dir{\sect}}(\cOne_A)$} node[below] {$\cOne_{\sect} \leftarrow 0 $} (B)

            (B) edge  [bend right=10] node[above, yshift=-0.15cm] { $\begin{array}{l}  \cOne_{ \sect} = \sectDist \end{array}$} node[below] {}  (C)
            (C) edge  [bend left=30] node[above] { $\sync_{\dir{\sect}}(\cOne_{B})$} node[below, yshift=0.05cm] {$\cOne_{\sect} \leftarrow 0 $}  (B)
            (C) edge  [bend right=10] node[above] { $\sync_{\dir{\sect}}(\cOne_A)$} node[below, yshift=0.05cm] {$\cOne_{\sect} \leftarrow 0 $}  (B)
            (C) edge  [bend right=25] node[above] {$\cOne_{ \dir{\sect}} = \sectL + \sectDist$}  (A);

    \end{tikzpicture}
    \caption{The intersection automaton of $ \sect $ where $ A$ and $ B$ can drive in direction $ \dirTrue $.}
    \label{fig:ta-inter}
\end{figure}
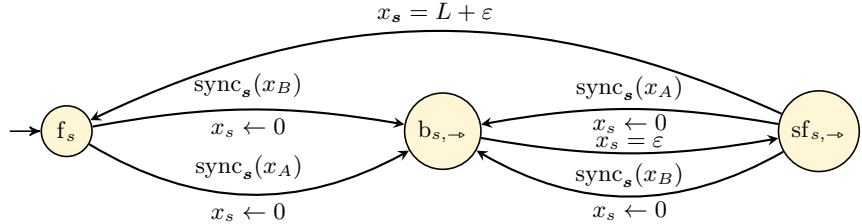

If a section $\sect$ is indeed an intersection,
then multiple copies of its sub-automaton will be present in the automata of the
corresponding cars. In order to ensure that a trace in this abstract model allows for collision avoidance in the real-world, we need to synchronize the different copies. To this end we create the intersection automaton shown in fig \ref{fig:ta-inter} for two cars sharing a section in direction $\dirTrue$.

The three states correspond to:
\textbf{free} $ (\free_{\sect} $): any car can enter in any direction;
\textbf{blocked} ($ \blocked_{\sect, \dirTrue} $): no car can enter;
\textbf{semi-free} ($ \semifree_{\sect, \dirFalse }$): an additional car can enter in $\dirTrue$ direction.
Together with the car automaton it ensures that at least $\epsilon$ time units pass
before two cars can enter the same section in the same direction, respecting the security distance in an abstract fashion. Similarly, two cars entering the section in opposite direction must be separated by at least $\sectL + \epsilon$
time units, allowing the first car to cross the section safely before letting
the second car in, as shown in the appendix \ref{apdx:automata}, page \pageref*{apdx:automata}, in fig \ref{fig:ex-inter-ta-two-ways}.
Note that all guards only involve equality testing, which means we could additionally
reset the clock to the same value, making our SWA effectively ISWA.

This ensures the safety distance, however not the relative order between the cars:
To prevent cars from reversing their order in $\wait_{\dir{\sect'}}$,
cars have to announce themselves on the successor section by pushing their token on
the corresponding channel $\channels_{\dir{\sect''}}$ before entering $\arrived_{\dir{\sect'}}$.
By doing so, the transition from waiting to driving is only activated if the
relative order is respected.

\subsection{Computing the optimal strategy for reachability}
\label{subsec:ta/algo-lvl-1}
Several mature tools handling timed automata able to answer reachability problems like
IMITATOR (\cite{andre2021}),
Uppaal (\cite{behrmann2006}, \cite{behrmann2007})
Tchecker (\cite{tchecker}) exist.
However, at the time of our experiments, none of them supported all the features we needed:
channels, stopwatches and time optimality.

What would be difficult for a general approach is to detect that a state cannot lead to a solution faster than the best one found so far, and it would typically lead to unnecessarily large computation times. This is what motivated us to propose a specific algorithm for time optimal reachability dedicated to our context.
\subsubsection*{A time optimal reachability algorithm}
Our goal is to obtain a time optimal trace witnessing reachability and use it later to construct our control algorithm.
We propose a Depth First Search (DFS) forward exploration that uses the properties of our systems of ISWA, synchronized with channels.
As we do not use the full expressiveness of parameters, as defined in Parametric Timed Automata (\cite{hune2002linear}, \cite{andre2016decision} and \cite{andre2019s})
(in our model, they never appear in any guards nor are they reset)
we can compare the obtained traces and store the current best trace that minimizes the global time.
Moreover, we can prune the states during the exploration using a conservative heuristic, leading to significant
performance gains.

The full description of our algorithm is presented in appendix \ref{apdx:algorithm}. Here, we give a very brief overview of its principle.
States consist of the configuration (location, zone representation via Difference Bounded Matrix (DBM)) and the channels' configuration.
We extend the subset relation for zones, denoted $\preceq$, to bounded channels: let $ s, s' $ be two states with respective zone $ z, z'$,
$ s \sqsubseteq s'$ iff they have the same location, the same channel configuration
and $z \preceq z'$ holds. This gives a partial order on states. As all our guards consist in checking equality, we can compute the successors by letting time elapse in all locations, until a clock matches the equality constraints of the future transition.
Finally, we use the conservative heuristic to detect if the current state implies a necessarily larger global time, in which case it is discarded.

\section{Ensuring feasibility of high-level plans}
\label{sec:smt}

Our first layer guarantees correctness
in the abstract setting, but is in general not physically realistic/realizable.
This gap between the physical reality and the high-level plan has to be
closed, or at least bounded, in order to obtain an actually feasible plan.
We 
extract \textit{important} events and their relative
order, which represent the \textit{combinatorial} part of the problem,
and retain it for the refined solution.
We so to speak built upon the high level plan to obtain a more realistic
sequence of inputs which still guarantees correctness.

In our running example, the high-level plan is represented by the time
optimal SWA-trace.
Recall that the control input, the current speed of the car,
is represented by the derivative of the clock associated to it.
In particular this entails that the car can only be stopped (derivative equals 0)
and the car drives at nominal speed (derivative equals 1) and
there is no time-delay between them corresponding to infinite acceleration,
a hardly realistic assumption.

To provide a more realistic model, we rely on satisfiability modulo theories (SMT).
The SMT framework provides great flexibility and expressiveness resulting from
the wide range of theories and functionalities disposable:
Quantifiers, Theory of reals, Minimization of an objective function, arrays and functions etc.

For our running example, using these advanced
functionalities comes at a very high cost and we therefore avoid them.
We rely on a discretization to allow for a good trade-off between
model accuracy and the complexity of the resulting problem.
Moreover, instead of minimizing the global time using the
built-in minimization functionalities of z3, it has proven to be more effective
to perform a linear search, solving the problem for some
given maximal global time repeatedly.

\subsection*{From SWA traces to piecewise linear control laws}

The SMT instance for traffic control is composed of one set of constraints
directly derived from the hybrid system and a second set of constraints representing
the high-level plan.
We model the traffic system as each car having a piecewise constant
speed and discretize time into $ \stepsNb$ steps of equal duration, that we denote $ \duration $.
We create a variable representing the speed of the
$i$th car during the $k$th time step, denoted $\dSpeed{i}{k}$, and denote $\dPos{i}{k}$ the position,
in its path, of the car $i$ at the beginning of the
$k$th timestep.
Without loss of generality, we suppose that $\duration = 1$.

We derive two constraints to guarantee physical realizability under a bounded error.
The position is the integral over the velocities, a simple sum as the
velocities are piecewise constant: $\dPos{i}{k} = \sum_{l=0}^{k-1} \dSpeed{i}{l} $.
Secondly we need to bound acceleration and velocity:
\begin{align*}
    \forall i,\ \forall k \in [0  \cdots \stepsNb - 2],  \ & \ \dSpeed{i}{k} - \maxDecc \le  \dSpeed{i}{k+1} \le \dSpeed{i}{k} + \maxAcc \label{eq:maximal-acc-decc} \\
    \forall i,\ \forall k \in [0  \cdots \stepsNb - 1], \ & \  0 \le \dSpeed{i}{k} \le \maxDSpeed 
\end{align*}
where $\maxDecc$ (\resp $ \maxAcc $) is the maximal deceleration (\resp acceleration) and
$\maxDSpeed$ is the maximal speed of the car ensuring a smaller gap between
the actual capabilities of the car and the ones implied by the resulting
reference trajectory.

To ensure coherence between the abstract solution represented by the SWA trace
and the refined solution represented by the piecewise constant speed,
as well as to reduce the search space of the SMT variables, we extract several
conditions. Let us denote by $\dRelPos{i}{\sect}{0}$ (\resp $\dRelPos{i}{\sect}{1}$) the position at
which the $i$th car \textbf{enters} (\resp \textbf{leaves}) some section $ \sect $, which is
obviously only defined if the total trajectory of the $i$th car passes
through section $ \sect $.

\textbf{Relative event order}
We want to ensure that \textit{important}
events happen in the same chronological order as given by the SWA trace.
\textit{Important} events, in our running example, are the moments when
cars enter or leave a section. We impose the constraints for each
pair of car and event.
Suppose car $i$ enters (\resp leaves) the section $\sect$ before
car $j$ enters (\resp leaves) section $\sect'$, this can be imposed by:
\begin{equation*}
    \forall k \in [ 0 \cdots \stepsNb - 2], \forall \switch \in \{ 0, 1 \}, \dPos{i}{k} < \dRelPos{i}{\sect}{\switch} \implies \dPos{j}{k} < \dRelPos{j}{\sect'}{\switch} %\label{eq:relative-event-order}
\end{equation*}

\textbf{Safety distance}
Whenever a car uses an \textit{intersection},
we need to ensure that it respects a security distance, denoted $ \security $, from the other cars,
even if the two cars are not currently sharing a section.
Suppose $\car_i$ enters section $ \sect $ before $\car_j$,
then for any $ k \in [0 \cdots \stepsNb - 2 ] $:
\begin{equation*}
    (\dRelPos{i}{\sect}{0} \le \dPos{i}{k} \le \dRelPos{i}{\sect}{1} \wedge
     \dRelPos{j}{\sect}{0} \le \dPos{j}{k} \le \dRelPos{j}{\sect}{1}) \implies
    (( \dPos{i}{k} - \dRelPos{i}{\sect}{0}) -( \dPos{j}{k} - \dRelPos{j}{\sect}{0}) > \security)
\end{equation*}

\textbf{Approximate timing}
To further restrict the search space for the
SMT problem and to increase the similarity between the abstract and refined solution,
we do not only keep the relative order between \textit{important} events,
but we also impose that they happen at approximately the same global time.

To do this, we introduce a parameter $\param$,
which limits the difference between the global time at which an \textit{important} event happens
in the high-level plan and the refined plan.
This permits a trade-off between the similarity of the
high-level, the refined plan and the danger of the SMT instance becoming unsatisfiable
(smaller value for $\param$ implies higher similarity however the high-level plan
may be infeasible for the more realistic model rendering the SMT instance unsatisfiable).

More formally, consider the important event of $\car_i$ entering
$\sect$ at the global time $\globaltime_0$. To ensure that the event will actually happen at most $ \param $ time-units later, we impose:
$
    \dPos{i}{\globaltime_0 + \param} \ge \dRelPos{i}{\sect}{0},
$
since the duration of each step equals $1$ (w.l.o.g.).

\textbf{Interpreting the solution}
If a satisfying solution for the SMT instance is found,
we can readily extract the refined plan from it.
All information necessary are the speed values for each time-step and car, \ie the value for all the $\dSpeed{i}{k}$.

\section{Reinforcement learning to get real-time policies}
\label{sec:RL}
Our global approach is divided into 3 stages, and in the previous sections we have presented the 2 first stages, which correspond to distinct levels of abstraction of the multi-agent system we want to control.
The solver presented in the two first stages in section \ref{sec:ta} and \ref{sec:smt} are collectivelly called SMT-SWA solver.
Given initial conditions of the system, these two stages enable us to get trajectories for all agents that solve the problem, but not in real-time, so if new initial conditions are faced at a high frequency, and if a high responsiveness is required, then the approach is not practical.
For the third stage of our layered approach, which we present in this section, we create a dataset of SWA-SMT solutions on a large number of random instances of the problem, and use this dataset as the initial experience replay buffer of a reinforcement learning (RL) algorithm.
Thereby we will first obtain a policy that imitates and slightly generalizes the SWA-SMT solutions, and will then progressively improve. At the end of the learning process, we get a neural network policy that can react in real-time to new conditions and can control the multi-agent system with a high success rate. 
We could also try to directly solve the multi-agent control problem with RL, with an initially empty experience replay buffer, but with our running example we show that for complex problems, the SWA-SMT solutions are crucial: without the initial dataset, the RL algorithm fails to find any solution to the problem, while with the initial dataset, the RL algorithm quickly matches and then outperforms the success rate of the SWA-SMT approach. In fact, RL algorithms are efficient at progressively improving control policies based on a dense reward signal, but problems with both continuous and combinatorial aspects may result in rewards that are very difficult to find. Multi-agent systems often have these properties, and the associated hard exploration problems are well known failure cases for standard reinforcement learning algorithms \cite{NIPS2016_afda3322}. We believe that in this context, a layered approach as the one presented in this paper can be particularly efficient. Using high level abstractions and formal verification, we ignore most of the continuous aspects of the problem, but solve its the most combinatorial and discrete parts, and get traces that can be refined into acceptable solutions. We then build a dataset that can be exploited by reinforcement learning to quickly get good policies, and then iteratively improve them.

To apply reinforcement learning, we cast the problem as a Markov Decision Process (MDP) with a state space $S$, an action space $A$, an initial state distribution $p(s_0 \in S)$, a reward function $r(s_t \in S, a_t \in A, s_{t+1} \in S)$ and transition dynamics $p(s_{t+1} \in S |s_t \in S, a_t \in A)$. Since the running example we consider is deterministic, we more specifically use a deterministic transition function: $s_{t+1} = \texttt{step}(s_t, a_t)$.
In this MDP, valid SWA-SMT trace should be directly interpretable as high reward episodes. See appendix~\ref{apdx:mdp} for a detailed description of the elements of the MDP for our running example.

Using the initial state distribution, we define random instances of the problem, and run the SWA-SMT solver to get valid solutions, i.e. traces.
We then transform each trace into an episode of the MDP.
To do so, we first retrieve the sequence of states and actions (see appendix~\ref{apdx:mdp} for details), then 
we compute the reward for all transitions of the episode, and we terminate the episode if a terminal state is reached (which happens only at the end because we only consider successful traces). The episodes we get correspond exactly to episodes of the MDP previously defined (again, see appendix~\ref{apdx:mdp} for details). We should discard episodes exceeding the maximum number of transitions (85), but we have set this number conservatively so that the time optimal SWA-SMT solutions are always shorter than the limit.

For our running example, we used the SWA-SMT solver on random initial states to generate 2749 successful episodes (with reward $\geq$ 2000, see appendix~\ref{apdx:mdp}) resulting in a total number of 176913 transitions. About 15\% of the random initial conditions were solved by the SWA-SMT solver and led to successful traces.\footnote{
   Generating a successful SWA-SMT trace takes on average about 15sec on a
   Intel i5-1235u with 16GB of RAM. Note that there is a high variance in this
   runtime ranging from under a second to several minutes. A timeout was set to
   900sec.
}

\begin{figure}
   \begin{center}
      \includegraphics[width=0.9\linewidth]{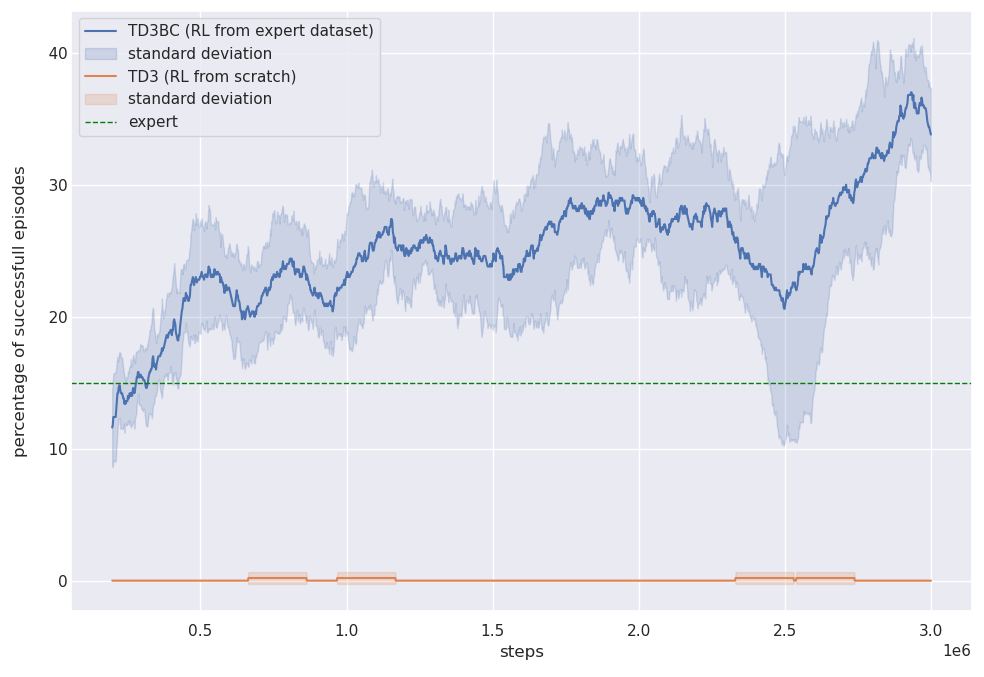}
      \caption{Percentage of successful episodes during training.}
      % RL from scratch is not able to solve the problem, but TD3BC, with the replay buffer initially filled with 176913 transitions obtained from successful episodes generated by the SWA-SMT solver, is quickly able to reach a success rate similar to the one of the SWA-SMT solver ($\sim 15$\%), and then improves over the training to reach $\sim 35$\% after 3 million steps.
      \label{fig:rl}
   \end{center}
\end{figure}

For the reinforcement learning, we select off-policy algorithms \cite{precup2001off} that use a replay buffer to store experience (episode transitions). During training, random batches of transitions are sampled from the buffer, and gradients of loss functions computed on these batches are used to iteratively update the parameters of neural networks (typically the policy network or actor and the value network or critic). New episodes are continuously run with the trained policy to fill the buffer. We compare two approaches, one in which an RL algorithm is trained from scratch (with an initially empty replay buffer), and one in which an RL algorithm starts with its replay buffer filled with the 176913 transitions collected from the SWA-SMT solutions.
More specifically, to perform RL form scratch, we first use TD3 \cite{fujimoto2018addressing}, a popular off-policy reinforcement learning algorithm, whereas TD3BC \cite{fujimoto2021minimalist} is used to perform RL with the replay buffer.
TD3BC is TD3 with a slight modification: in the actor loss, a regularization term of behavioral cloning is added, helping the RL algorithm to handle and imitate expert training data that does not come from the trained policy.
Originally designed for offline RL (\ie RL on purely offline data, without running episodes),
TD3BC can also be seen as a variant of TD3 that is able to start with a non-empty replay buffer initialized with expert data.
We use exactly the same hyperparameters for TD3 and TD3BC (see appendix~\ref{apdx:params}),
and for the additional behavioral cloning regularization term in TD3BC,
we use the default parameter $\alpha = 2.5$ (cf. \cite{fujimoto2021minimalist}).
Fig \ref{fig:rl} show the results we obtained with a training of 3 million steps on 5 random seeds for each method\footnote{Using the \emph{xpag} RL library \cite{xpag},
 with a single Intel Core i7 CPU,
  32GB of RAM, and a single NVIDIA Quadro P3000 GPU,
   the training took between 40 and 50 minutes per million steps.}.
   We define successful episodes as episodes with a cumulated reward greater than 2000,
   which only happens when each car reaches its final destination.
   We observe that the first approach (TD3 from scratch) never learns to solve the problem.
   On the other hand, after 250k steps (one step is one discrete time step in an episode play with the neural network policy being trained),
   the second approach (TD3BC) already reaches the same success rate as the SWA-SMT solver (about $15$\%),
   and then it continues to improve during the 3 million steps of training.
   In the end, we obtain neural network policies with a success rate of approximately $35$\% in average,
   which is more than twice the success rate of the SWA-SMT solver.
   So we not only obtain policies that can be executed in real-time,
   we also obtain policies that find solutions more consistently.

\begin{figure}[h]
\begin{center}
   \includegraphics[width=0.95\linewidth]{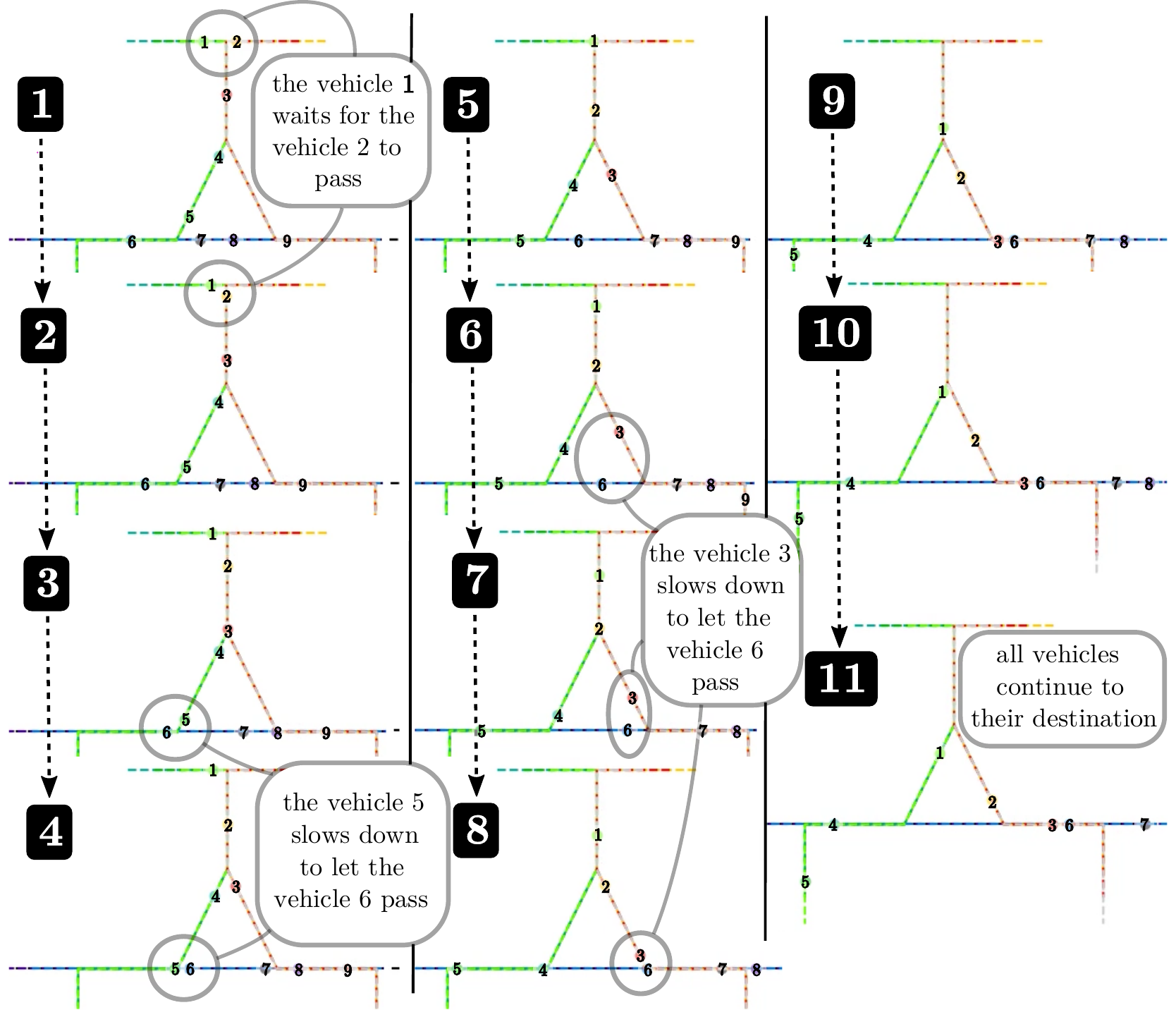}
   \caption{A successful episode played by a policy trained with TD3BC.}
   \label{fig:example}
\end{center}
\end{figure}

However, while the SWA-SMT solutions are optimal by construction
(\ie they achieve success with the minimum number of steps),
there is no such guarantee with the neural network policies trained via reinforcement learning.
Fig~\ref{fig:example} shows an episode played by a neural network policy trained with TD3BC. A full video of this episode and a few others is hosted at \url{perso.eleves.ens-rennes.fr/people/Emily.Clement/Implementation/multi-agent.html}
The tool we implemented in open-source and can be found at \url{gitlab.com/Millly/robotic-synthesis}.

\section{Conclusion and future work}
\label{sec:conclu}
We presented a layered approach for multi-agent control involving three stages,
the two first ones relying on formal verification tools to compute time optimal solutions, and the last one relying on these solutions (called the SWA-SMT data) to guide a reinforcement learning algorithm.
We demonstrated the effectiveness of the approach by applying it to a centralized traffic control problem,
showing that,
thanks to the SWA-SMT data, the RL algorithm quickly learns to solve the problem and progressively improves to obtain higher success rates.
Our results demonstrate the potential of combining layered approaches with RL for multi-agent control.
The high-level abstraction in the first stage places emphasis on the combinatorial elements of the problem, leading to high-level plans which are then refined into solutions addressing the actual continuous dynamics of the agents. While it is difficult to construct these solutions in real-time, a rich enough dataset of such solutions can be used to efficiently guide the reinforcement learning, thus eliminating the need for the RL algorithm to tackle the difficult task of exploring the combinatorial aspects of the multi-agent control problem.
Ultimately, we obtain a neural network policy capable of controlling the multi-agent system in real-time.
In future work, we would like to implement our proposed algorithm for time optimal reachability in ISWA with bounded channels in the open-source tool TChecker~\cite{tchecker},
and apply our layered approach to decentralized multi-agent systems.

\section*{References}
\printbibliography[heading=none]

\clearpage
\appendix

\section{Description of the systems of Timed Automata of for car traffic control}
\label{apdx:automata}

\subsection*{An example}
Let us first fully describe the intersection automata of section $ \sect $, presented in section \ref{subsec:ta/application} if we add a car, called C, where the directed section $ ( \sect, \dirFalse ) $ appears in its path.
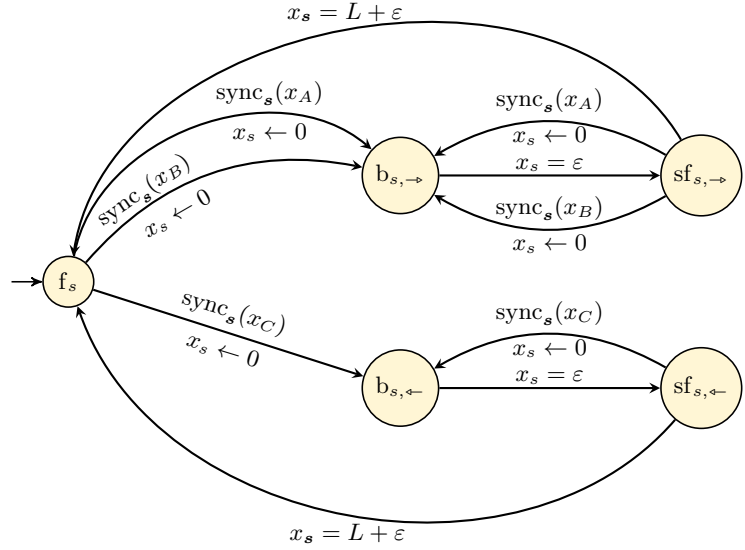
\begin{figure}[htb]
    \centering
    \begin{tikzpicture}[->,>=stealth',auto,node distance=2cm, semithick]
        \node [initial state] (A) [] {$\free_{\sect}$};
        \node [state] (B) [above right of=A, xshift=3cm] {$\blocked_{\sect, \dirTrue}$};
        \node [state] (C) [right of=B, xshift=2cm] {$\semifree_{\sect, \dirTrue}$};

        \node [state] (D) [below right of=A, xshift=3cm] {$\blocked_{\sect, \dirFalse}$};
        \node [state] (E) [right of=D, xshift=2cm] {$\semifree_{\sect, \dirFalse}$};

        \path[fleche]
            (A)  edge [bend left] node[above, auto=left, rotate=35, yshift=0.05cm] {$\sync_{\dir{\sect}}(\cOne_{B})$} node[pos=0.18,auto=right, rotate=30] {$\cOne_{\sect} \leftarrow 0 $} (B)

            (A)  edge [bend left=60] node[above, auto=left, pos=0.7] {$\sync_{\dir{\sect}}(\cOne_A)$} node[pos=0.7,auto=right] {$\cOne_{\sect} \leftarrow 0 $} (B)
            (B) edge  [left] node[above, yshift=-0.15cm] { $\begin{array}{l}  \cOne_{ \sect} = \sectDist \end{array}$} node[below] {}  (C)
            (C) edge  [bend left] node[above] { $\sync_{\dir{\sect}}(\cOne_{B})$} node[below, yshift=0.05cm] {$\cOne_{\sect} \leftarrow 0 $}  (B)
            (C) edge  [bend right] node[above] { $\sync_{\dir{\sect}}(\cOne_A)$} node[below, yshift=0.05cm] {$\cOne_{\sect} \leftarrow 0 $}  (B)
            (C) edge  [bend right=70] node[above] {$\cOne_{ \dir{\sect}} = \sectL + \sectDist$}  (A);
        \path[fleche]

            (A)  edge [left] node[above, pos=0.5, rotate=-17] {$\sync_{\dir{\sect}}(\cOne_C)$} node[below, pos=0.5, rotate=-17] {$\cOne_{\sect} \leftarrow 0 $} (D)
            (D) edge  [left] node[above, yshift=-0.15cm] { $\begin{array}{l}  \cOne_{ \sect} = \sectDist \end{array}$} node[below] {}  (E)
            (E) edge  [bend right] node[above] { $\sync_{\dir{\sect}}(\cOne_C)$} node[below, yshift=0.05cm] {$\cOne_{\sect} \leftarrow 0 $}  (D)
            (E) edge  [bend left=60] node[below] {$\cOne_{ \dir{\sect}} = \sectL + \sectDist$}  (A);

    \end{tikzpicture}
    \caption{An intersection automaton of a section $ \sect $ such that $ A$ and $ B$ can drive in $ \sect $ in direction $ \dirTrue $ and $ C$ can drive in direction $ \dirFalse $.}
    \label{fig:ex-inter-ta-two-ways}
\end{figure}

\subsection*{Construction of the system of ISWA}

We have two types of timed automata in our system: automata representing cars and automata representing intersections.
\subsubsection*{Clocks and parameter}

We use a parameter, which never appears in any guard, invariant or reset, to represent the \textit{global time}, we denote it $ t $.
To describe the progression of each car, we assign, for each car $ \car $, a clock, denoted $ \clock_{\car} $ representing $\car$'s position along its path.
We also assign a clock for each intersection $ \sect $ that we denote $\clock_{\sect}$. $ \clock_{\sect} $ is common for both direction.
This clock represents the following value:
the value of $ \relativePosition{ \car }{ t }{ \dir{\sect} } $, where $ \car $ is the car that is closest to the beginning of the section (whatever direction is used).

\subsubsection*{Channels}

To report the presence of a car in a section, each section is assigned a channel.
Each time a car wants to enter a section, we push its name as symbol in the channel before entering it, where entering it means accessing the associated waiting state.
To be able to drive within the section, we must be able to read its symbol in the channel, that means that this car is the first entered the section among all entered cars.
This ensures that a car does not overtake another car.

\subsubsection*{Automata of cars}
Let $ \car= ( \sect_k, \direction_k )_{ 0 \leq k \leq m} $ a car of $ \setOfCars $. We assigned a clock, denoted $ \clock_{\car} $ to this car.
\begin{description}
    \item[\textbullet\,Locations:] for each directed section $ \dir{\sect} $, a car can either \textit{drive} in $ \dir{\sect} $, \textit{have reached the end} of $ \dir{\sect} $ and \textit{wait} in $ \dir{\sect} $.
    In the last case, we model the waiting state by assigning a stopwatch for $ \clock_{\car} $.
    \item[\textbullet\,Transitions:] a car can change from waiting to driving in the same section, from driving to arriving and finally from arriving to waiting (to the next section).
    1) From waiting to driving, the condition to respect is that $ \clock_{\car} $ is equal to the value of the $ \clock_{\car} $ when entering in $ \dir{\sect} $.
    Additionally, we need to respect the order constraint imposed by the channel.
    2) From driving to arriving, the clock $ \clock_{\car} $ must have increased by $ \sectL $ (length of the section $ \sect $)
    and we push the name of the car in the channel dedicated to this section.
    3) From arriving to waiting in the next section, the clock must not have increased and need to synchronize with the intersection automaton if the successor section is an intersection.
\end{description}

\subsubsection*{Automata of intersection section}

To drive in a section, we have to check beforehand that our car will not collide with another in an abstract section.
As each speed is equal between the cars, we can easily check this.
For this, the clock associated to the section, $ \clock_{ \sect } $ is reset each time a new car enters the intersection.

The automaton of $ \sect $ is composed of three or five locations, depending on whether the cars can travel in one or both directions:
\begin{description}
    \item[\textbullet\,Blocked Location:] no cars can enter.
    \item[\textbullet\,Semi-free location:] a (unique) car can enter, if it is in the same direction.
    \item[\textbullet\,Free location:] any car can enter, regardless of its direction.
\end{description}
From a Blocked location, we must allow time, equal to the safety distance, to elapse. When this amount of time has elapse, we pass a transition from Blocked to Semi-free location, with the same direction.

Again, from a semi-free location, we have two choices:
either another car (with the same direction) enters, reseting the clock and we arrive in a blocked location (with the same direction),
or we let time elapse (equals to the length addition with the security distance) and we arrive in Free location.

To indicate a car $ \car $ enters, the action from semi-free (or free) to blocked, is a synchronized action with the one activated in the automaton of car $ \car $, when $ \car $ enters in the intersection section.
Therefore, each automaton of an intersection section has synchronized actions with all cars that will eventually enter in this section.

\section{Reachability algorithm for ISWA synchronized with channels}
\label{apdx:algorithm}
Our Depth First Search based algorithm allows us to obtain a time optimal trace for reachability.
To do so, we store the fastest trace found so far and compare candidates to this
solution during the exploration.
Let us describe precisely our algorithm.

We begin by initializing several variables:
\begin{itemize}
    \item a set of \textit{explored states}, called \texttt{explored}.
    It stores all the states the algorithm has explored yet, in order to avoid exploring a state twice. It is initialized as the empty set.
    \item \texttt{bestSol}$=(\texttt{bestSolTime}, \texttt{bestSolTrace})$ a tuple of the total time spent to reach the goal, \texttt{bestSolTime}, and the trace of the run, \texttt{bestSolTrace}.
    It represents the best solution found at each step of the algorithm.
    It is therefore naturally initialized at $ ( + \infty, \texttt{None} ) $ so that any first trace found be better than it.
    \item An initial state, $s_0$ which is a tuple of all the initial configurations (locations and valuations) of the Timed Automata of our system of ISWA.
    \item a stack, denoted $ \texttt{stack} $, which stores pairs of state and the possible successors of this state.
    The stack is initialized with $ s_0 $ and the iterator of the successors of $ s_0 $.
\end{itemize}
In our algorithm, we explore the future states, computed by the function \texttt{succ}.
If and when we reach the goal state, \ie when all the goal locations of our system of timed automata are reached, we compare the trace of this run with the last stored trace.
If the global time is smaller, we replace the stored trace by the trace of our current run.

A conservative heuristic is used to avoid exploring states guaranteed to produce traces
that are not time optimal.
\SetKwComment{Comment}{/* }{ */}
% \begin{algorithm}[htb]
% \caption{Time optimal reachability}\label{alg:OptReach}

%     $explored\gets \{\}$\Comment*[r]{Set of explored states}
%     $bestSol\gets (\infty, None)$\Comment*[r]{Best solution so far}

%     $s_0\gets init()$\Comment*[r]{Initial state}
%     $stack\gets ()$\Comment*[r]{Stack pair(state, successors iterator)}
%     $stack.push((s_0, SuccessorIterator(s_0)))$\;

%     \While{len(stack)}{
%         $nextState \gets stack.head()[1].next()$\;
%         \If{$nextState$ is $None$}{
%             $stack.pop()$\;
%             $continue$\;
%         }
%         \If{$isFinal(nextState)$}{
%             \If{$minGlobTime(nextState) < bestSol[0]$}{
%                 $bestSol \gets (minGlobTime(nextState), traceOf(stack, nextState))$\;
%             }
%             $continue$\Comment*[r]{Global time can only grow $\rightarrow$ Ignore successors}
%         }
%         \If{$\exists\ s \in explored\ s.t.\ nextState \sqsubseteq s$}{
%             $continue$\;
%         }
%         \If{not $keepExploring(nextState)$}{
%             $continue$\Comment*[r]{Conservative heuristic}
%         }
%         $stack.push((nextState, SuccessorIterator(nextState)))$\;
%     }
%     $return\ bestSol$\;
% \end{algorithm}

\begin{algorithm}[htb]
    \caption{Time optimal reachability}\label{alg:OptReach}
        $explored\gets \{\}$\Comment*[r]{Set of explored states}
        $bestSolTrace\gets None$\Comment*[r]{Best solution so far}
        $bestSolTime\gets \infty$\Comment*[r]{Time of bestSol}

        $s_0\gets init()$\Comment*[r]{Initial state}
        $stack\gets ()$\Comment*[r]{Stack pair(state, successors iterator)}
        $stack.push((s_0, succ(s_0, None)))$

        \While{stack is not empty}{
            $currState, currSucc \gets stack.pop()$\;
            $nextState \gets succ(currState, currSucc)$\;
            \If{$nextState$ is $None$}{
                $continue$ \Comment*[r]{No more successors}
            }
            $stack.push((currState, nextState))$\;
            \If{$isFinal(nextState)$}{
                \If{$minGlobTime(nextState) < bestSolTime$}{
                    $bestSolTrace \gets traceOf(stack, nextState)$\;
                    $bestSolTime \gets minGlobTime(nextState)$
                }
                $continue$\Comment*[r]{Global time can only grow $\rightarrow$ Ignore successors}
            }
            \If{$\exists\ s \in explored\ s.t.\ nextState \sqsubseteq s$}{
                $continue$\Comment*[r]{The subsequent state has already been explored}
            }
            \If{not $keepExploring(nextState)$}{
                $continue$\Comment*[r]{Conservative heuristic}
            }
            $stack.push((nextState, None))$\;
        }
        $return\ bestSol$\;
    \end{algorithm}

Let us detail some very useful sub-functions of our algorithm \ref{alg:OptReach}.

\subsubsection*{\textbullet\,isFinal:}
the helper function \texttt{isFinal} determines whether the target location has been reached.

\subsubsection*{\textbullet\,minGlobTime:}
the helper function \texttt{minGlobTime} extracts the minimal value for the global time with which the target can be reached from the zone.

Since we are only interested in the fastest trace and the global time is never
stopped nor reset, we do not need to visit the successor states of an accepting state.

\subsubsection*{\textbullet\,keepExploring:}
To speed up the algorithm, we use the function \texttt{keepExploring}
as a heuristic to determine whether it is worth to keep exploring the currently
state.
In order to ensure correctness of the algorithm, this function needs to be
conservative. Meaning that it may never discard a state for which,
if the exploration had continued, an accepting trace faster than the currently best
solution could be found.

In our running example, this function sums the current global time plus the maximal
difference between any current car position and its respective goal position.
Intuitively this function computes the time it would take for all cars to arrive
if collisions can be ignored from the current situation on.
Naturally this heuristic is therefore conservative.

\subsubsection*{\textbullet\,succ:}
The \texttt{succ} function computes, given a current state \texttt{currState} and the associated successor iterator \texttt{currSucc}, the next successor to explore, denoted \texttt{nextState}.

%A tricky part could be to compute the time elapsing between a state and its successor.
To compute it, we use the special properties of our model.

In our system, described in appendix \ref{apdx:automata}, the constraints of
our timed automata only contain equality tests of the form $ \clock == \clock_0 $ where $ \clock $ is a clock of the system of timed automata and $ \clock_0 $ is a constant.

Therefore, we do not need to represent zones, but simple valuations contain all information necessary;
to compute the delay that elapses before taking the next transition, we just have to compute the minimal delay between all the possibilities.
%  knowing the minimal delay before any transition in our system is taken.
%  our current state, the minimal delay elapsing before a transition is taken in a timed automaton in the system.

 More precisely, we have these following situations:
 \begin{itemize}
    \item A transition in a car-automaton can be taken.

    If this car is in a waiting location, there are two possibilities: the car takes the transition or it waits until at least another transition is taken in the whole system.

    Otherwise, the transition is taken, as there is no stopwatch.
    \item In an intersection-automaton, we can be in these three type of locations:

    \begin{enumerate}
        \item If the location is a Free location, then the choice depends on the car-automaton.

        \item If the location is a Blocked Location, the transition is taken and it \enquote{semi-frees} the intersection, as described in appendix \ref{apdx:automata}.

        \item If the location is a Semi-free location, two choices are possible: another car asks to enter or the time has elaspe enough to pass to a Free location.
    \end{enumerate}
 \end{itemize}

\section{Markov Decision Process for the running example}
\label{apdx:mdp}
The Markov Decision Process is defined by its state space $S$, action space $A$, initial state distribution $p(s_0 \in S)$, reward function $r(s_t \in S, a_t \in A, s_{t+1} \in S)$ and deterministic transition function $s_{t+1} = \texttt{step}(s_t, a_t)$.

We describe here all the elements of the MDP defined for our running example:

\begin{itemize}
\item \textbf{The state space ($\mathbb{R}^{720}$).} The environment contains 3 paths, which are unions of sections, and as detailed in section~\ref{sec:Appli}, we have imposed a maximum number of 3 cars per path, so there are at most 9 cars, to which we can attribute a unique identifier (we use $\{-1, 0, 1\}^2$). The state is entirely defined by the speed and position of each car. We could thus use vectors of size 18 to represent states, but instead we chose a sparser representation with a better structure. To remain coherent, we use the same road network presented in sec. \ref{sec:Appli} which is composed of 24 section (since every car has a dedicated initial and goal node subdividing the sections containing initial and goal positions).
On this road network, three different paths are defined, and each section being shared by at most 2 paths.
At any given time any section may contain at most 6 cars by construction. For each section, we define a list of 6 tuples, all equal to (0, 0, (0, 0), 0) if no car is currently inside the section. However if there are cars in the section, say 2 cars for example, then the first two tuples have this structure:
$$
(\texttt{position with the section},
\texttt{normalized velocity},
\texttt{car identifier},
1)
$$

We represent states as a concatenation of the values of all these tuples for all the 24 sections, which amounts to a vector of size 720. It is a sparse representation, but its advantage is that it makes it easy to find cars close to each other, as they are either in the same section or in neighbor sections.
\item \textbf{The action space ($\mathbb{R}^9$) and transition dynamics.} Given an ordering of the 9 cars, an action is simply a vector of 9 accelerations. If $a_i$ is the acceleration for the car $i$, and if at the current time step its position within its path is $p_i$, and its speed is $v_i$, then at the next time step its position will be $p_i + v_i$, and its speed will be $v_i + a_i$. This defines the transition dynamics of the MDP. The components of an action corresponding to cars that are not present in the state are simply ignored. Remark: actions can be computed straightforwardly from a sequence of states as they are equal to the difference between consecutive speeds for each car.
\item \textbf{The reward.} When all cars have reached their destination, i.e. crossed the end of their path, a reward of 2000 is given, and the episode is terminated. Besides, when there is either a collision (a violation of the safety distance between two cars) or two car facing each other in opposite directions in the same section, a negative reward (-100) is given and the episode is terminated. Finally, at each time step, two positive rewards are given, one proportional to the average velocity of the cars (to encourage cars to go fast), and one proportional to the (clamped) minimum distance between all cars (to encourage cars to stay far from each other). We set the maximum number of time step per episode to 85, and adjust these rewards so that an episode cannot reach a cumulated reward of 2000 unless it is truly successful and gets the final +2000 reward.
\item \textbf{The initial state distribution.} We define an arbitrary initial state distribution in which each of the 9 cars has an 80\% chance of being present. The speed of each car is defined randomly, and positions are also defined randomly (within roughly the first two third of each path). Safety distances are ensured, so that the inital states are not in collision, however speeds may be such that there will a collision after the first time step, so there is no guarantee of feasibility.

\end{itemize}

\section{Hyperparmeters of the RL algorithms}
\label{apdx:params}
For TD3:

\begin{itemize}
\item Actor network architecture: multi-layer perceptron (MLP) \cite{almeida1997c1} with 3 hidden layers of size 256 and rectified linear unit (ReLU) activation functions.
\item Actor optimizer: ADAM \cite{2015kingma}, actor learning rate: $10^{-3}$
\item Critic network architecture: MLP with 3 hidden layers of size 256 and ReLU activation functions.
\item Critic optimizer: ADAM, critic learning rate: $10^{-3}$
\item Discount factor: $0.99$
\item Soft update coefficient ($\tau$): $0.05$
\end{itemize}

For TD3BC: 

\begin{itemize}
\item Actor network architecture: MLP with 3 hidden layers of size 256 and ReLU activation functions.
\item Actor optimizer: ADAM , actor learning rate: $10^{-3}$
\item Critic network architecture: MLP with 3 hidden layers of size 256 and ReLU activation functions.
\item Critic optimizer: ADAM, critic learning rate: $10^{-3}$
\item Discount factor: $0.99$
\item Soft update coefficient ($\tau$): $0.05$
\item $\alpha$: $2.5$
\end{itemize}

\end{document}